\documentclass{article}
\usepackage{spconf,amsmath,graphicx}
\usepackage{flushend}
\usepackage{algorithm}
\usepackage{algorithmicx}
\usepackage{mathtools}
\usepackage{algpseudocode}
\usepackage{subfigure}
\usepackage{multirow}
\usepackage{longtable}
\usepackage{amsthm}
\usepackage{amssymb }
\usepackage{amsmath,tabularx}
\usepackage{bm}
\usepackage{mathrsfs}
\usepackage{amsfonts}
\usepackage{longtable}
\usepackage{multirow}
\usepackage{cite}
\usepackage{color}
\usepackage{graphicx}

\def\RLF{\text{RL-IFF}}

\def\mA{\mathcal{A}}
\def\IoTD{\text{IoT-TD}}

\def\mR{\mathcal{R}}

\def\k{_{k}}
\def\nk{_{k+1}}

\def\w{\bm{w}}

\def\s{\bm{s}}

\def\x{\bm{x}}

\def\y{\bm{y}}

\title{Hybrid Indoor Localization via Reinforcement Learning-based Information Fusion }
\ninept

\name{Mohammad Salimibeni$^\dag$, Arash Mohammadi$^\dag$}
\address{$^\dag$~Concordia Institute of Information Systems Engineering (CIISE), Montreal, Canada}

\begin{document}

\maketitle

\begin{abstract}
The paper is motivated by the importance of the Smart Cities (SC) concept for future management of global urbanization. Among all Internet of Things (IoT)-based communication technologies, Bluetooth Low Energy (BLE) plays a vital role in city-wide decision making and services. Extreme fluctuations of the Received Signal Strength Indicator (RSSI), however, prevent this technology from being a reliable solution with acceptable accuracy in the dynamic indoor tracking/localization approaches for ever-changing SC environments. The latest version of the BLE v.$5$.$1$ introduced a better possibility for tracking users by utilizing the direction finding approaches based on the Angle of Arrival (AoA), which is more reliable. There are still some fundamental issues remaining to be addressed. Existing works mainly focus on implementing stand-alone models overlooking potentials fusion strategies. The paper addresses this gap and proposes a novel Reinforcement Learning (RL)-based information fusion framework ($\RLF$) by coupling AoA with RSSI-based particle filtering and Inertial Measurement Unit (IMU)-based Pedestrian Dead Reckoning (PDR) frameworks. The proposed $\RLF$ solution is evaluated through a comprehensive set of experiments illustrating superior performance compared to its counterparts.
\end{abstract}
\begin{keywords}
Angle of Arrival (AoA), Bluetooth Low Energy (BLE), Convolutional Neural Network (CNN), Indoor Localization, Internet of Things (IoT).
\end{keywords}
\maketitle
\vspace{-.15in}
\section{Introduction} \label{sec:introduction}
\vspace{-.1in}
It is expected that in near future Smart Cities (SCs)~\cite{SalimiICASSP2021} will flourish across the globe aiming at efficient management and control of global urbanization to assist in having more reliable and safe societies. Internet of Things (IoT)~\cite{SalimiICASSP2021}, plays a vital role in developing different SC services by providing an interactive infrastructure. The large volume of data provided by IoT devices can assist in providing services to control and solve the different SC problems, especially in indoor environments. The key underlying challenge here is to efficiently fuse and process such large amounts of IoT data. Reinforcement Learning (RL) techniques~\cite{Parvin:access} are attractive solutions for this challenge (to fuse IoT data) as RL approaches can learn and adapt to the environmental changes and can directly learn from historical SC data where having a unified model is too expensive to develop.
Among different SC services, Contact Tracing (CT)~\cite{Xu2021, Salimibeni2021} for pandemic control has become of paramount importance especially given recent COVID-19 pandemic. The main building block of autonomous CT frameworks is the proximity/localization module. Existing CT models are, primarily built for outdoors and can not offer scalable solutions for indoor environments. On the other hand, existing indoor localization techniques, typically, utilize a single technology or are developed based on a single processing model. Therefore, such solutions commonly suffer from different issues such as sensitivity to multi-path effects, noise, fluctuations of received signal, and frequency/phase shifts. Capitalizing on issues associated with stand-alone solutions and the importance of the indoor localization topic, especially during this pandemic area, the paper focuses on multi-modal and Reinforcement Learning (RL)-based information fusion techniques that allow for simultaneous integration of different technologies and processing solutions.

\noindent
\textbf{Literature Review:}
To achieve a reliable localization service, one major approach is to leverage the information from multi-sensor systems to improve the system robustness, accuracy of the prediction and enhance the detection range. Information fusion is one major approach in multi-sensors IoT networks which can be categorized into traditional methods and Machine Learning (ML)-based~\cite{Meng:2020, Himeur:2022} approaches. Probabilistic fusion, evidential belief reasoning fusion, fuzzy theory, and tensor fusion are among the main traditional information fusion strategies~\cite{Federico:2013}.
In~\cite{Meng:2020}, practical use of ML methods as an information fusion technique is reviewed and different information fusion levels including signal-level, feature-level and decision-level are discussed. Data labeling and training to fuse data obtained from multiple data sources, however, is a time consuming and costly procedure~\cite{Saha:2020, Li:2020}.
RL-based methods~\cite{Guo:2022, Han:2022, Chen:2020, Zhou:2020} can address such issues for resource-constrained IoT edge devices. Generally speaking, RL targets providing human-level adaptive behavior by construction of an optimal control policy~\cite{Sutton2018}. The main underlying objective is learning (via trial and error) from previous interactions of an autonomous agent and its surrounding environment~\cite{Salimibeni:2022, salimiICASSP, Parvin:access, Spano, Seo, Salimibeni2020Asi}.
For example in~\cite{Guo:2022}, to address the problems related to the weighted fusion method with fixed weight allocation in decision level fusion, a deep RL multi-modal decision-making fusion weight allocation is developed. The powerful decision-making ability of deep RL is used in this approach to mitigate the problems related to the traditional fusion models. Liu \textit{et al.}~\cite{Liu:2022} proposed a deep RL multi-type data fusion framework to solve the issues related to the complicated stock market environment and fusing different data types for algorithmic trading services. In this approach, an static Markov Decision Process (MDP) is defined for the proposed signal fusion strategy and RL is mainly used when the main fusion task is done on temporal features of stock data and others to make trading decisions. The idea mainly is not following a RL-based fusion strategy and classical fusion solutions are applied in this work. Another priori knowledge RL-based information fusion method for multi-sensors for air combat data is proposed in~\cite{Zhou:2020}. In this paper, RL is used to find the coefficients for the fusion of the information for different sensory data input. An static MDP is also considered to be used in this approach to be able to apply RL on the fusion phase. Whereas some of the above-mentioned works may lack timeliness and are not applied for localization purposes, indoor localization tasks deal with a time-varying system. There are different sensor fusion localization approaches~\cite{Yu:2022, Sou:2022, Dinh:2021} proposed in literature, yet the fusion section is not implemented based on RL approaches, mostly relying on the traditional sensor fusion techniques or supervised ML techniques. Considering the problems related to the data labeling, the complexity of mathematical modeling of indoor environments, and time-varying nature of the information received from different sensor~\cite{Yu:2022} and achieved localization results from different approaches, the paper proposes an RL-based information fusion method to fuse the localization results gained from different localization modalities.

\vspace{.025in}
\noindent
\textbf{Contributions:}
To address the above mentioned challenges, we focus on development of a hybrid Received Signal Strength Indicator (RSSI)~\cite{Parvin:SPL}, Angle of Arrival (AoA)~\cite{Parastoo:Fusion}, and Pedestrian Dead Reckoning (PDR)~\cite{Amin:J} localization framework. In the context of hybrid BLE-based indoor localization, fluctuations of the received signal (related to the RSSI component); cumulative error in trajectory estimation (associated with the PDR component), and; the frequency/phase shifting between the transmitter and the receiver oscillators (corresponding to the AoA component) are the most prominent issues targeted. To handle these challenges and enhance the accuracy of the proposed localization/tracking framework, we fused the localization results of three employed approaches, i.e., RSSI, PDR, and AoA-based localization solutions. An innovative RL-based Fusion framework, referred to as $\RLF$, is utilized to deal with the time-varying nature of the information received from different localization solutions and non-stationary MDP.

The rest of the paper is organized as follows: In Section~\ref{Sec:Fus}, RL-based fusion strategy is described. Section~\ref{Sec:DFwPK} presents the proposed RL-based fusion strategy with Priori Knowledge. Section~\ref{sec:5} presents experimental results. Finally, Section~\ref{sec:6} concludes the~paper.
\section{RL-based Fusion Strategy} \label{Sec:Fus}
In this section, we leverage an RL-based fusion strategy to fuse the tracking estimates of RSSI, PDR and AoA-based localization solutions. The main objective is to find the optimal fusion weights, which is achieved by adjusting the RL learning procedure based on the following steps:

\noindent
$\bullet$ Based on the information related to the target received from multi sensor and a generated signal via the RL solution, an action is taken and the corresponding reward will be calculated.

\noindent
$\bullet$ The accumulative reward will be updated in each step taken in the environment. By having the new target information, an action will be chosen leveraging an action selection scheme.

\noindent
$\bullet$ Considering the time varying nature of the environment, an MDP is designed based on the fusion accuracy to address related challenges.

Each specific indoor environment has its own structure, dimensions, and static/time-varying environmental variables, therefore, model-based fusion strategies become increasingly complex,  implementation-wise. In an alternative approach, we use Q-learning, instead, to fuse the localization information. The goal of such a fusion task is to find the actions that collectively maximize the reward gained from the environment to reach optimality in fused values.  In each time step, the RL-based fusion engine (being at State $s_k$) selects an action from the Action set $\mA$ and moves to the next state $s_{k+1}$ receiving a scalar reward $r_k$ based on the reward function $\mR$. This learning scheme will continue via trial and error until reaching the terminal state. The Q-function is defined as follows
\begin{eqnarray}
 Q_{\pi}(\s, a) = \mathbb{E} \left\{\sum_{k=0}^{T}\gamma^k r\k |\s_0 = \s, a_0 = a, a\k = \pi(\s\k) \right\},\label{Eq:Q}
\end{eqnarray}
and the Q-learning is performed based on the following Bellman model
\begin{eqnarray}
\lefteqn{Q_{\pi^*}(\s\k, a\k) = Q_{\pi^*}(\s\k, a\k) +\nonumber}\\ && \alpha \Big(r\k +\gamma \max_{a\in\mA}Q_{\pi^*}(\s\nk, a)  - Q_{\pi^*}(\s\k, a\k) \Big).\label{Eq:4}
\end{eqnarray}
The synchronized tracking information gained from three tracking paths (i.e., RSSI, PDR and AoA) will be fused ($F_T$) based on a specific weight as follows
\begin{equation}\label{efuse}
\hat{\x}^{\text{Fused}}_k = \w^{\text{RSSI}}\x^{\text{RSSI}}\k +  \w^{\text{PDR}}\x^{\text{PDR}}\k +  \w^{\text{AoA}}\x^{\text{AoA}}\k,
\end{equation}
where $\x^{\text{RSSI}}\k$, $\x^{\text{PDR}}\k$, $\x^{\text{AoA}}\k$ are the tracking results in the time stamp $k$ and $\w^{\text{RSSI}}$, $\w^{\text{PDR}}$ and $\w^{\text{AoA}}$ are the associated fusion weights. These are computed based on our prior works~\cite{Parastoo:Fusion, Parvin:SPL, Amin:J}. In each time, summation of the weights is equal to one, i.e.,
\begin{equation}\label{efuse2}
\w^{\text{RSSI}} + \w^{\text{PDR}} + \w^{\text{AoA}} = 1.
\end{equation}
The RL solution is used in the fusion phase for finding optimal weights to fuse the results of three tracking scenarios.

\section{Data Fusion with Priori Knowledge}\label{Sec:DFwPK}
\setlength{\textfloatsep}{0pt}
\begin{figure} [t!]
\centering \includegraphics [scale = 0.15] {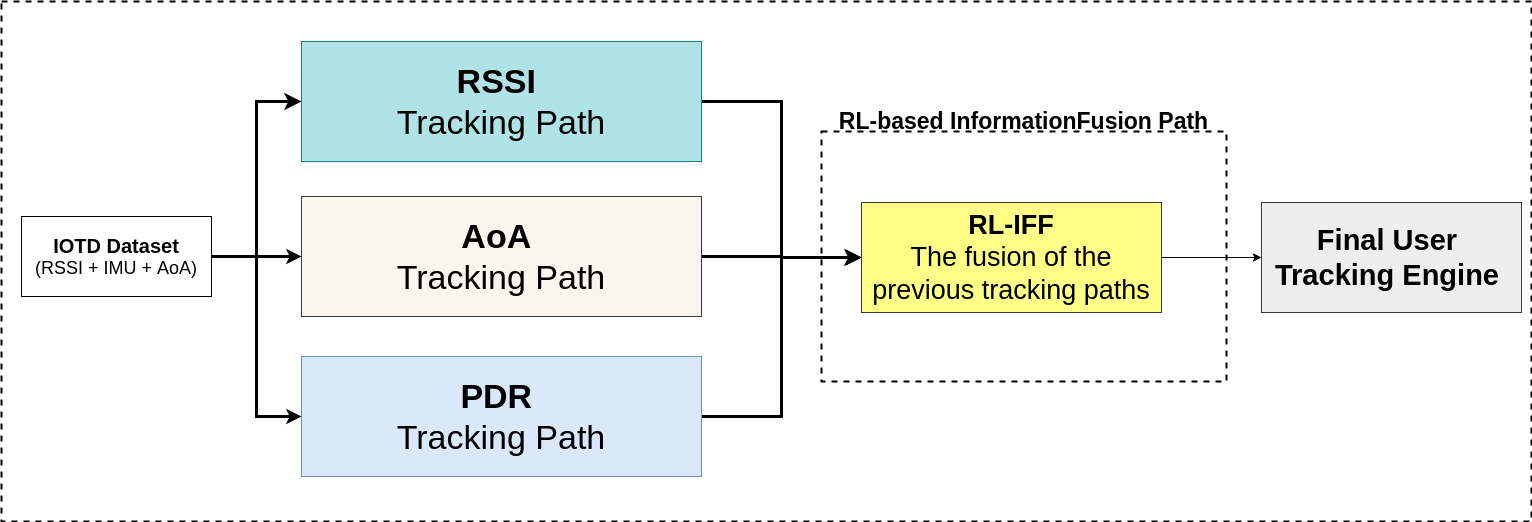}
\caption{\small The proposed Localization Fusion framework.} \label{fig:Fuse}
\end{figure}
\noindent
Considering the data gathered in different indoor locations and leveraging the $\IoTD$ dataset~\cite{SalimibeniEUSIPCO:2021}, we can have a priori knowledge of tracking scenarios in different indoor venues. Fig.~\ref{fig:Fuse} represents the overall structure of the proposed fusion framework. There are many challenges related to the non-stationary RL solutions~\cite{Hamadanian:2022}, specifically for the systems that are subjected to time-varying workloads. If the states in the tracking scenarios become defined based on the different physical locations or fingerprinting methods, the states would be continuous and hard to discretize. States' definitions also can be very wide considering the type of indoor environments and time-varying user location scenarios. Considering the aforementioned challenges, the proposed $\RLF$ is trying to deal with the time-varying MDP, and transform it into a static MDP definition. In this regard, the localization accuracy error is considered a key metric to define the states and the proportional rewards gained by the agent. In what follows, state, action and reward function definitions for the proposed RL solution are discussed thoroughly.
\vspace{-.15in}
\subsection{State, Action and Reward Functions of the $\RLF$}\label{Sec:DFwPK}
\vspace{-.1in}
Consider an agent in position $p_k$ ($p_k = (x_k, y_k)$), taking a step in an environment and moving to a new local position $p_{k+1}$($p_{k+1} = (x_{k+1}, y_{k+1})$). The tracking error can be the distance between the exact location of the user, i.e., $p_{(r, k)} = (x_{(r, k)}, y_{(r, k)})$, and the newly fused estimation of the location, i.e., $p_{(e, k)} = (x_{(e, k)}, y_{(e, k)})$. This error ratio is used to define the states and the reward gained by the agent. We assume $100$ states for this experiment, where the continuous error values are discretized. For the absolute error values between $0$ to $1$, we have $100$ states as follows
\begin{eqnarray}\label{Eq:state}
\lefteqn{\epsilon_k \!\!=\!\! \sqrt{(x_{(r, k)} -(\w^{\text{RSSI}}\x^{\text{RSSI}}\k + \w^{\text{PDR}}\x^{\text{PDR}}\k + \w^{\text{AoA}}\x^{\text{AoA}}\k))^2 +\nonumber}}\\
&&(y_{(r, k)} -(\w^{\text{RSSI}}\y^{\text{RSSI}}\k + \w^{\text{PDR}}\y^{\text{PDR}}\k + \w^{\text{AoA}}\y^{\text{AoA}}\k))^2,
\end{eqnarray}
\begin{eqnarray}\label{Eq:state2}
s\k = \begin{cases} round(\epsilon_k*100) , & \text{if } 0 \leq \epsilon_k < 1 \\ 100, & \text{if } \epsilon_k >= 1 \end{cases}.
\end{eqnarray}
For the absolute error values more than $1$, the state would be $100$.
Considering the above state definition, the reward function can be expressed as follows
\begin{eqnarray}\label{Eq:reward}
r\k = \begin{cases} 100, & \text{if } \epsilon_k = 0 \\ round(\frac{1}{round(\epsilon_k, 2)}) , & \text{if } 0 < \epsilon_k < 1 \\ -100, & \text{if } \epsilon_k >= 1 \end{cases}.
\end{eqnarray}
Considering Eq.~\eqref{Eq:reward}, the reward function $r\k$ is inversely proportional to the $\epsilon_k$. The smaller distance between the actual location of the user and the estimated location, the reward gained by the agent in the corresponding state would be larger.
A significant negative reward will be issued for the $\epsilon_k$ values equal to or bigger than $1$.
The proposed $\RLF$ aims to find the optimal weights to fuse the results of three tracking scenarios. Algorithm~\ref{algo:1}, briefly represents the proposed framework. Different actions taken in this platform can be defined based on weight adjustments. By modifying the weights, different fusion values can be calculated, and training can be performed based on the comparison between the new fused location estimation and the user's actual location in different states.
Since there are three tracking scenarios and their underlying weights, based on Eq.~\eqref{efuse2}, by assigning the new values to two of the weights, we can find the value of the third one. Consequently, the action function can be defined by $+$, $-$ and $<>$, denoting increased, decreased and unchanged operations on the weights, respectively as follows
\begin{eqnarray}\label{Eq:action}
\begin{cases} \w^{\text{x}} = (1+k\%)\w^{\text{x}} & \text{if } \w^{\text{x}} + \\ \w^{\text{x}} = (1-k\%)\w^{\text{x}} & \text{if } \w^{\text{x}} - \\ \w^{\text{x}} = \w^{\text{x}} & \text{if } \w^{\text{x}} <> \end{cases},
\end{eqnarray}
where $\text{x}$ is the tracking solution. Table~\ref{Table:tableAction}, shows the possible actions regarding the modification of the weights.
\begin{table}[t!]
\centering
\caption{\small Actions in the $\RLF$ for fusion strategy.}\label{Table:tableAction}
\begin{tabular}{| l | l |}
\hline
 \textbf{Action} & \textbf{Weight Modification} \\ \hline
1 & $\w^{\text{RSSI}} +$, $\w^{\text{AoA}} +$  \\ \hline
2 & $\w^{\text{RSSI}} +$, $\w^{\text{AoA}} - $\\ \hline
3 & $\w^{\text{RSSI}} +$, $\w^{\text{AoA}} <>$  \\ \hline
4 & $\w^{\text{RSSI}} -$, $\w^{\text{AoA}} -$  \\ \hline
5 & $\w^{\text{RSSI}} -$, $\w^{\text{AoA}} +$  \\ \hline
6 & $\w^{\text{RSSI}} -$, $\w^{\text{AoA}} <>$ \\ \hline
7 & $\w^{\text{RSSI}} <>$, $\w^{\text{AoA}} +$  \\ \hline
8 & $\w^{\text{RSSI}} <>$, $\w^{\text{AoA}} -$ \\ \hline
9 & $\w^{\text{RSSI}} <>$, $\w^{\text{AoA}} <>$ \\ \hline
\end{tabular}
\end{table}
\begin{algorithm}[t!]
\caption{\textproc{The Proposed RL-based Information Fusion Framework: $\RLF$}}
\label{algo:1}
\begin{algorithmic}[1]
\State \textbf{Learning Phase:}
\State \textbf{Input:} The tracking results of the applied tracking paths, i.e., the RSSI path; the PDR path, and; the AoA path;
\State \textbf{Output:} The optimal set of weights, fused data;
\State Initialize $Q = 0$ with random weights $\w^{\text{RSSI}}$,  $\w^{\text{AoA}}$ and  $\w^{\text{PDR}}$;
\State Set parameters $\gamma$ and $\alpha$ for Q-learning;
\State Initialize state $s_0$, leveraging the initial weights in Eq.~\eqref{Eq:state2};
\State  \textbf{Repeat for each episode}:
\State \quad $a\k \leftarrow$ epsilon greedy action selection in $s\k$
\State \quad Take action $a\k$, observe the next state $s\nk$;
\State \quad Calculate the gained reward using Eq.~\eqref{Eq:reward};
\State \quad ${\footnotesize Q(\s\k, a\k) \!\!\leftarrow\!\! (1-\alpha)Q(\s\k, a\k) \!\!+\!\!\alpha \Big(r\k \!+\!\gamma \underset{a\in\mA}{\arg\max}Q(\s\nk, a\k) \Big)}$;
\State \quad $\w^{\text{RSSI}}$,  $\w^{\text{AoA}}$ and  $\w^{\text{PDR}} \leftarrow$ optimal weights that maximize  $Q(\s\nk, a\k)$;
\State \quad $s\k \leftarrow$ optimal new state $s^*$;
\State \textbf{end for}
\State \textbf{return} $\w^{\text{RSSI}}$,  $\w^{\text{AoA}}$ and  $\w^{\text{PDR}}$; fused data $p\k$ .
\end{algorithmic}
\end{algorithm}
\begin{figure}[t!]
\centering
\includegraphics[width=4.1cm]{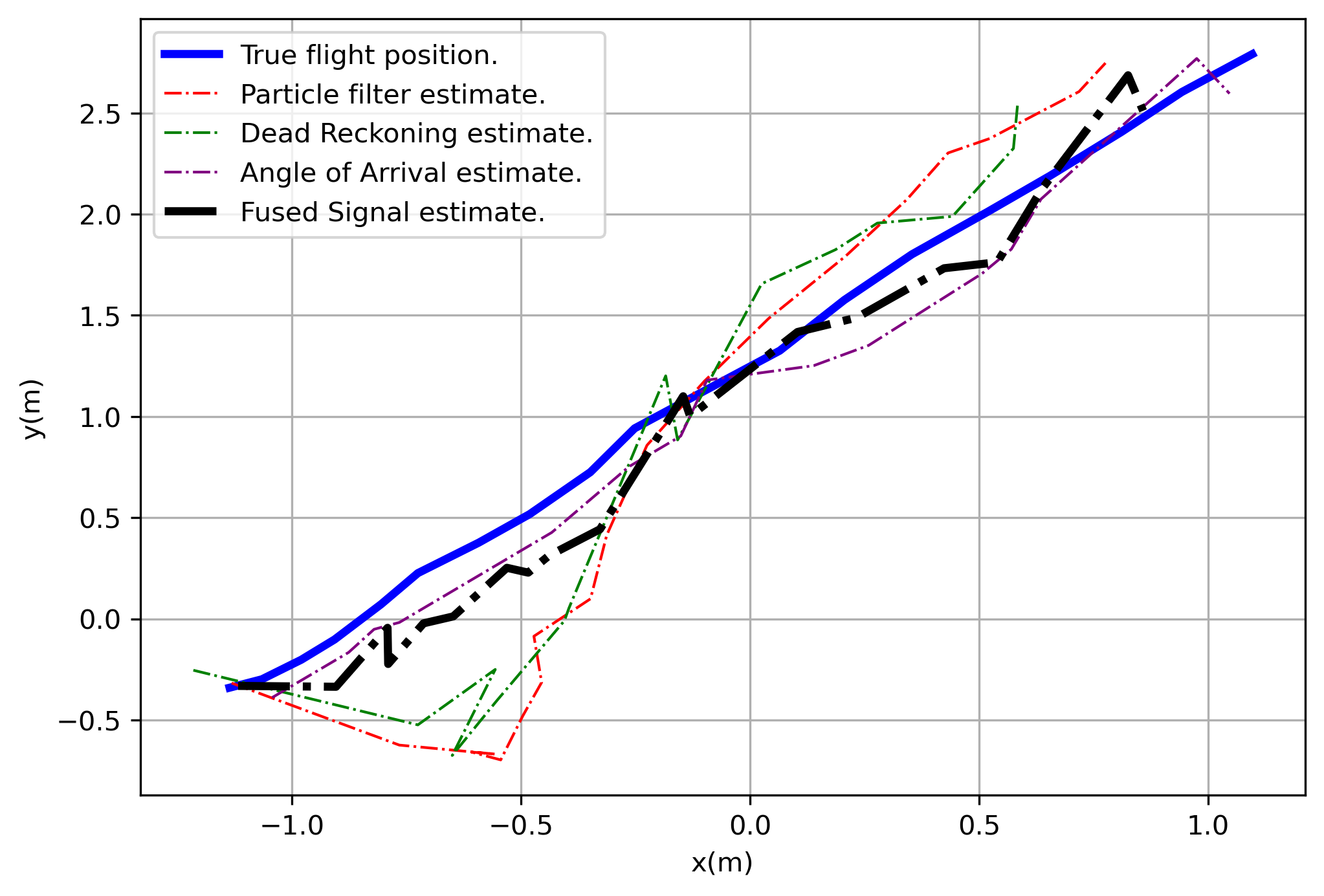}
\includegraphics[width=4.1cm]{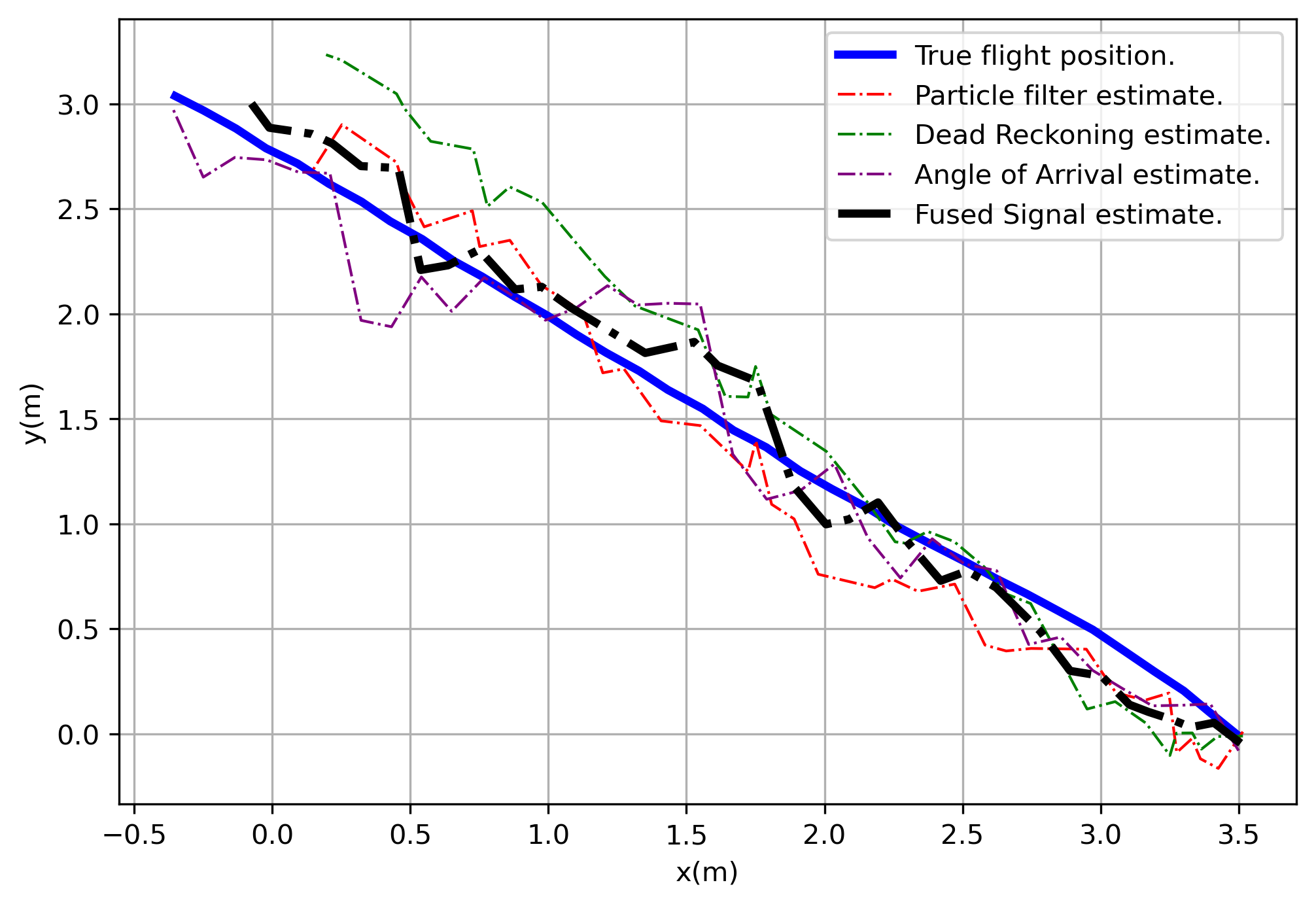}
\includegraphics[width=4.1cm]{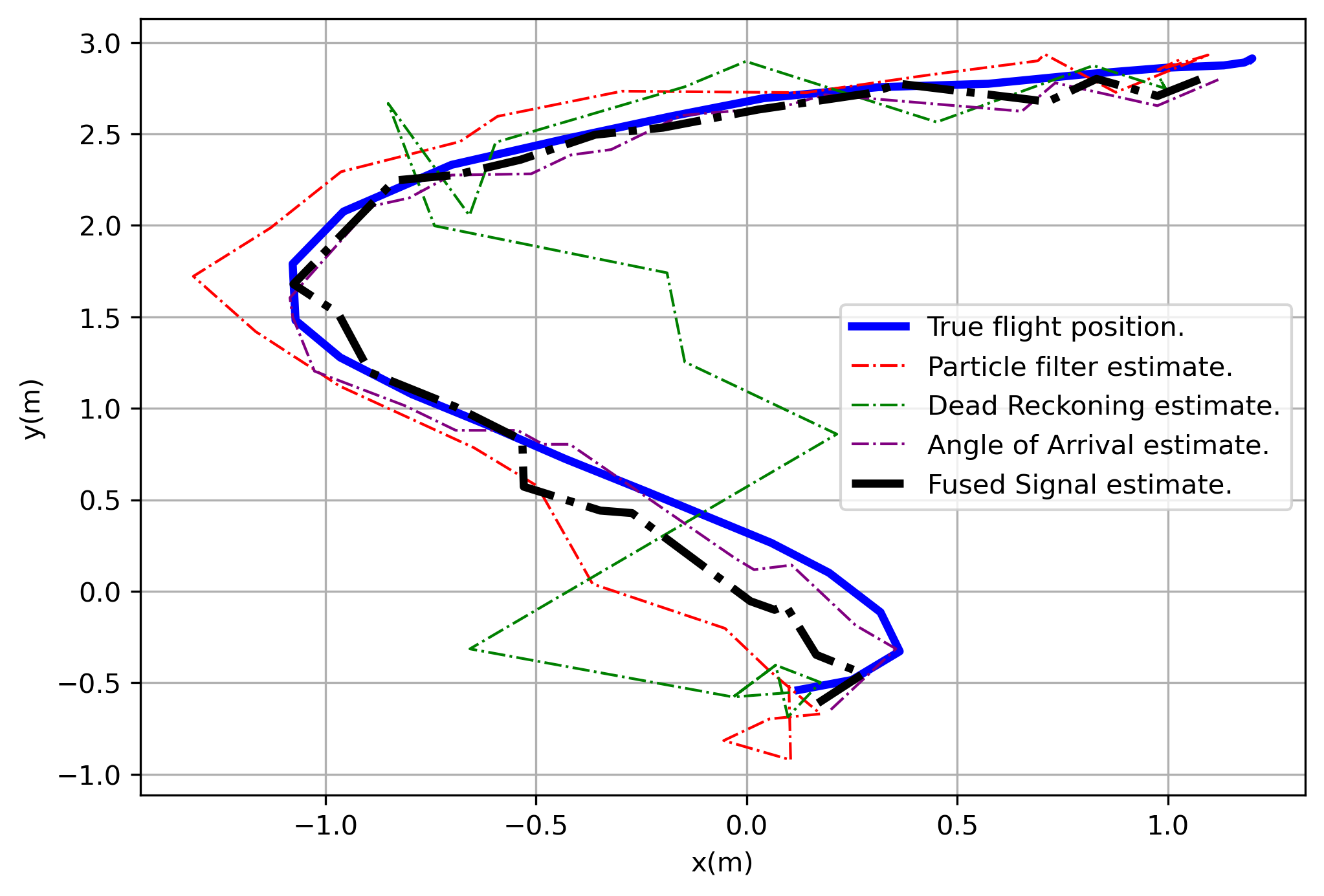}
\includegraphics[width=4.1cm]{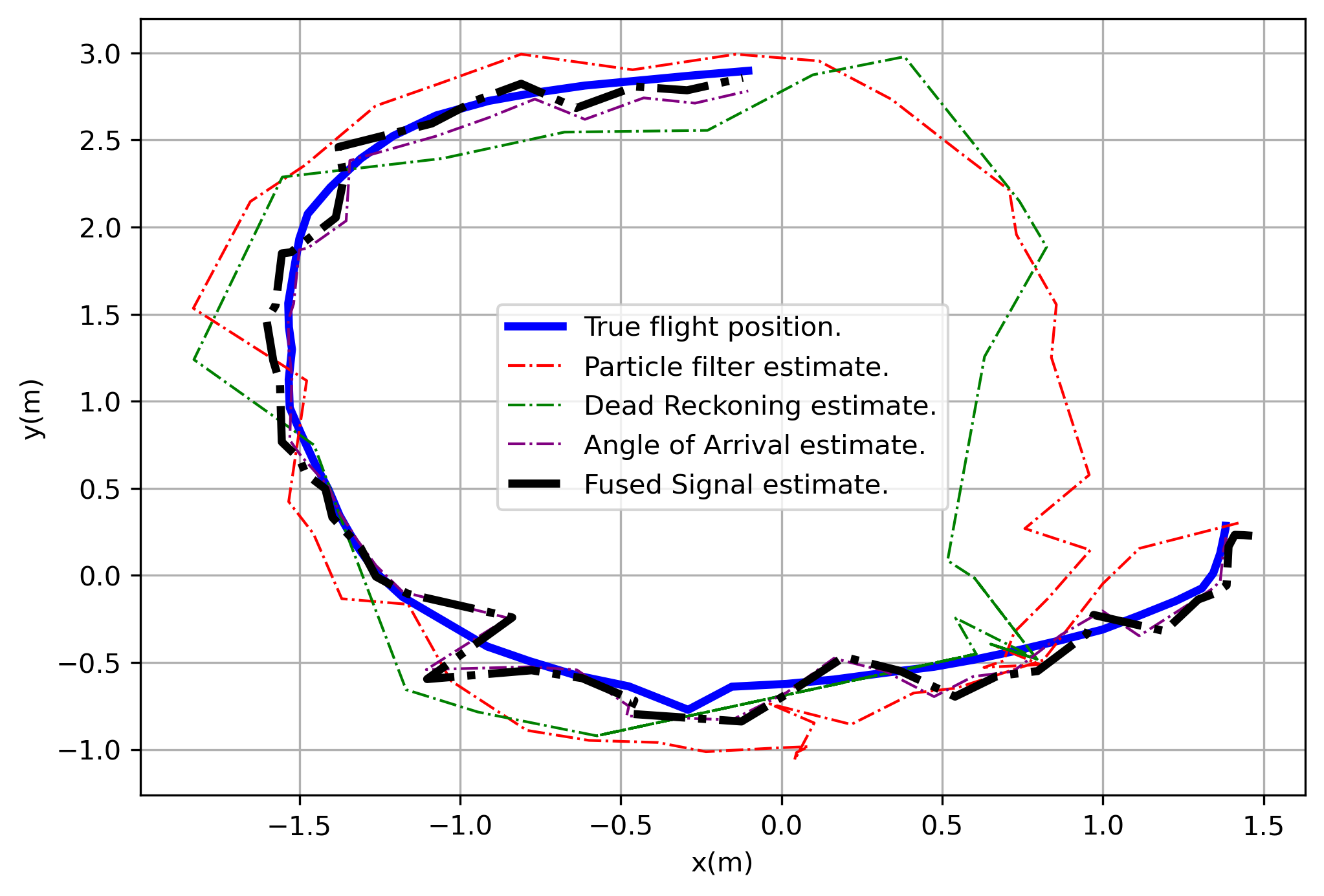}
\vspace{-.15in}
\caption{\footnotesize $\RLF$ Results for Four different movement scenarios in the second environment: (a) Diagonal A;  (b) Diagonal B; (c) Random, and (d) Rectangular trajectories.}\label{Fig:22}
\end{figure}
\begin{figure}[t!]
\centering
\includegraphics[width=4.1cm]{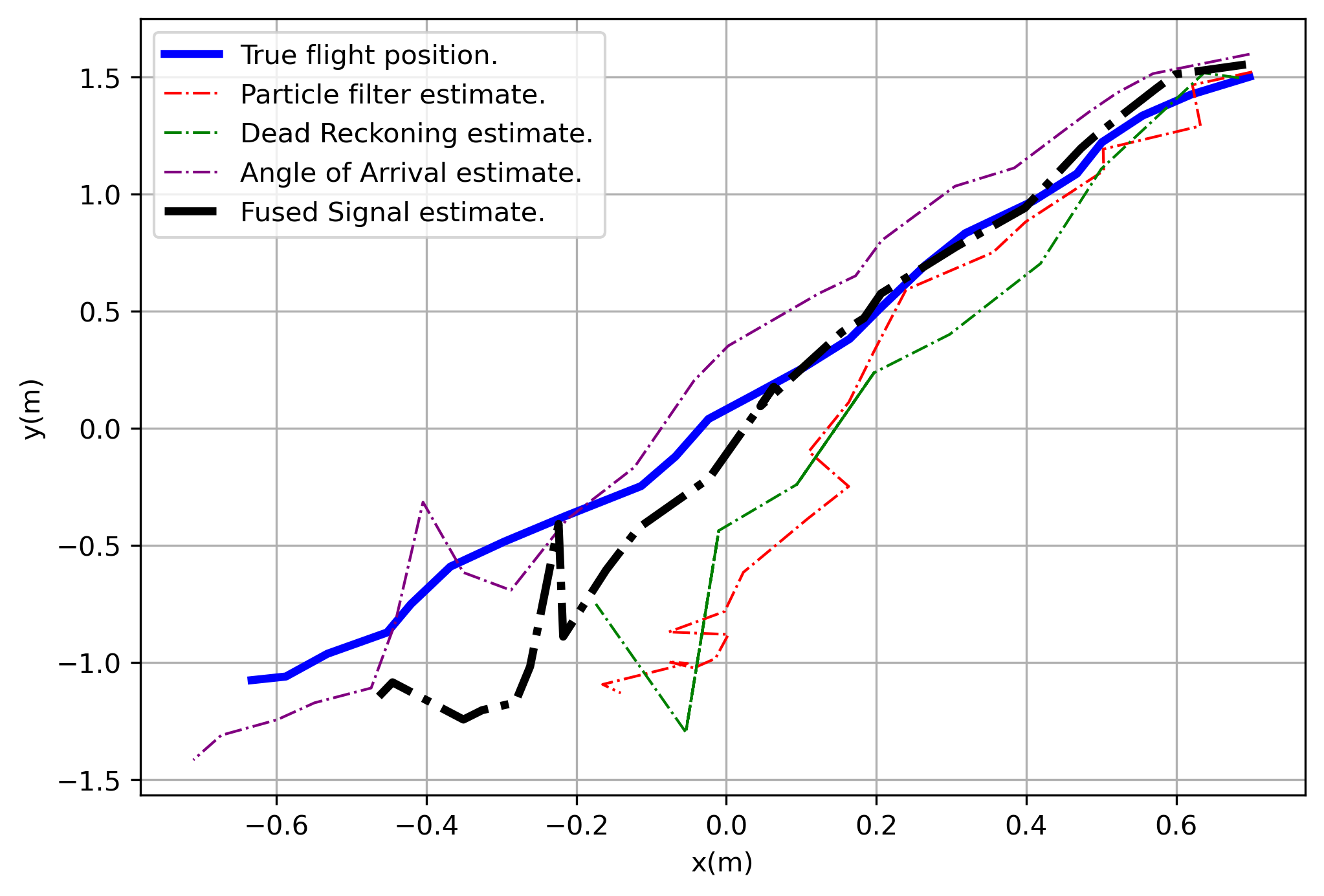}
\includegraphics[width=4.1cm]{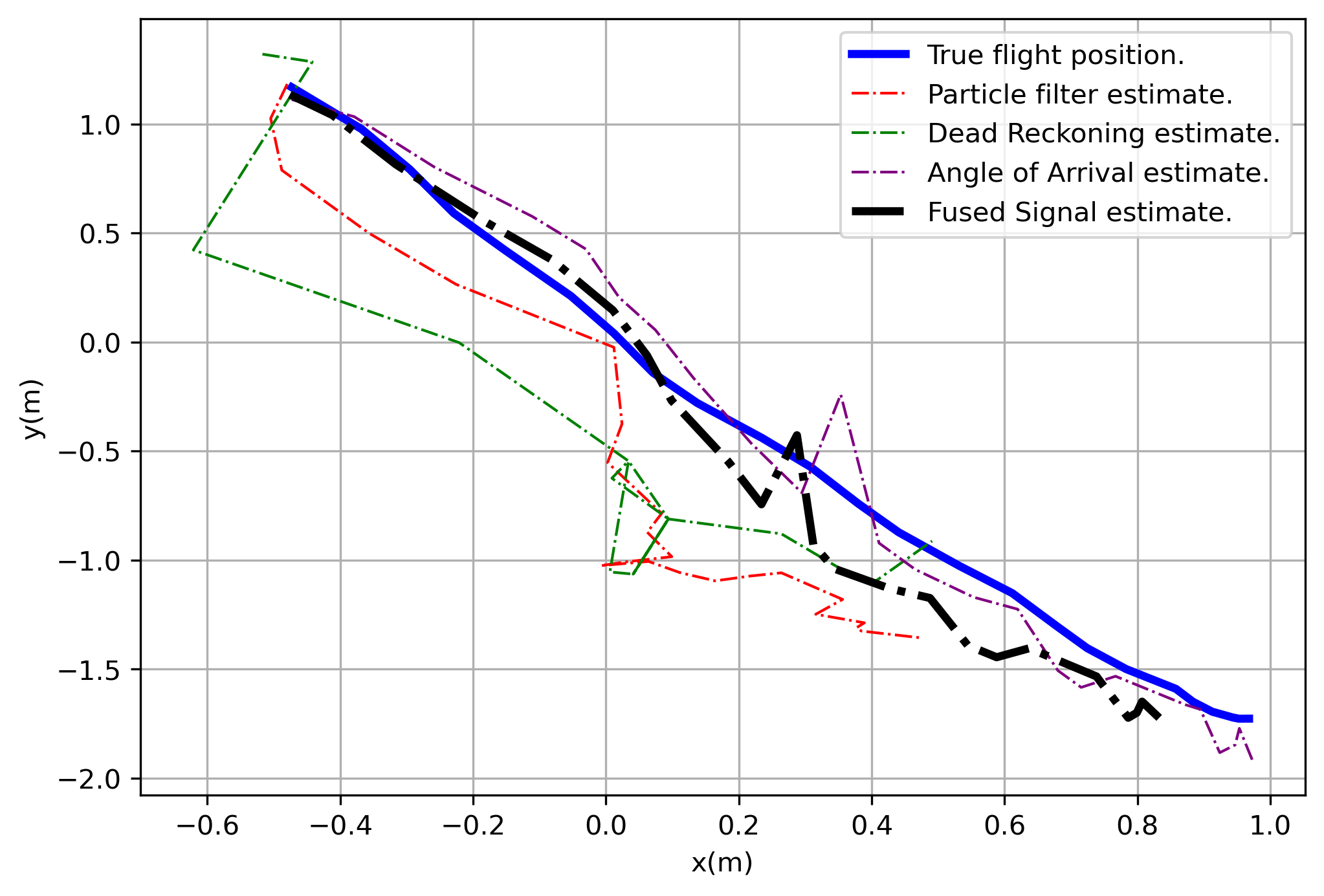}
\includegraphics[width=4.1cm]{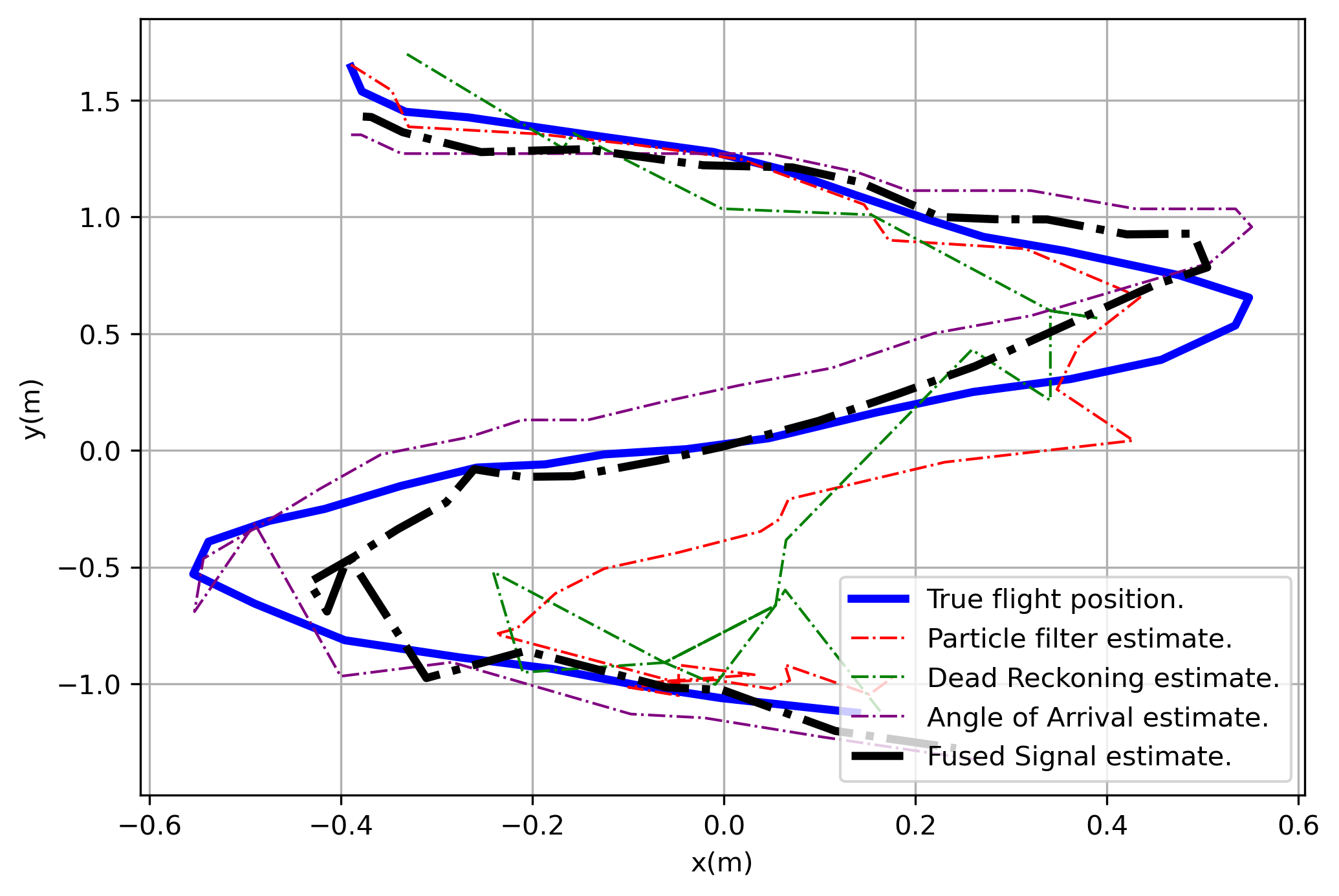}
\includegraphics[width=4.1cm]{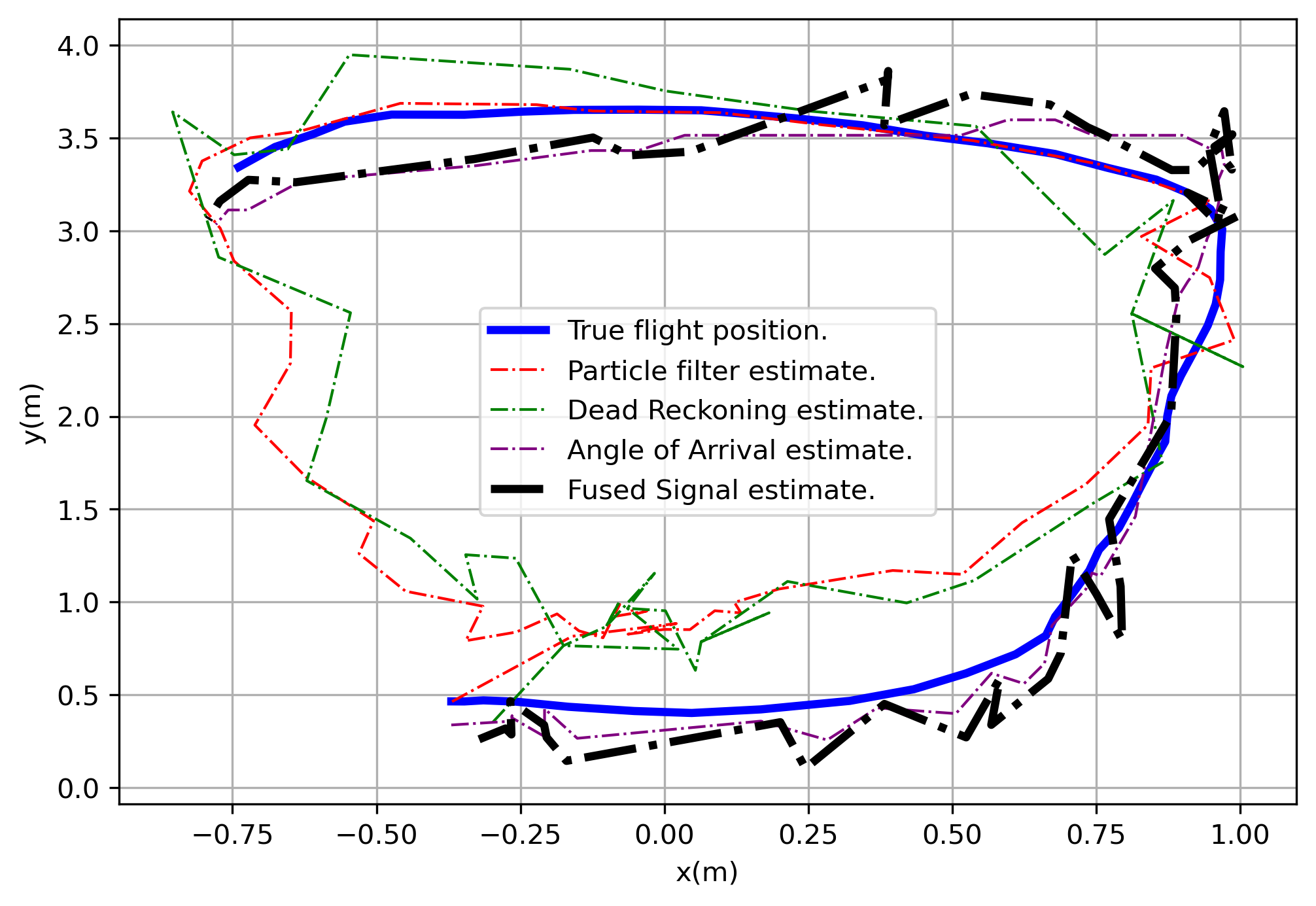}
\vspace{-.15in}
\caption{\footnotesize $\RLF$ Results for Four different movement scenarios in the second environment: (a) Diagonal A;  (b) Diagonal B; (c) Random, and (d) Rectangular trajectories.}\label{Fig:23}
\end{figure}
\vspace{-.15in}
\section{Simulation Results} \label{sec:5}
\vspace{-.1in}
In this section, we evaluate the $\RLF$ framework for different localization solutions leveraging the ground truth. The implemented localization techniques are based on the RSSI coupled with Kalman Filter (KF) and Particle Filter (PF) tracking algorithms, PDR-based tracking scheme~\cite{SalimibeniEUSIPCO:2021}, the AoA estimation algorithm~\cite{Parastoo:Fusion}, and the data is gathered in three different indoor locations by doing three different movement scenarios, i.e., rectangular, diagonal and random walking.
Figs.~\ref{Fig:22} and~\ref{Fig:23} show four different movement scenarios in two environments. In each environment, estimated trajectories are obtained based on the RSSI-based coupled with KF and PF tracking algorithm, PDR-based tracking scheme, the AoA estimation algorithm, the ground truth data, and the fused results of the tracking scenario. By applying the proposed $\RLF$ to find the optimal weights for information fusion, more accurate results are achieved.
%
\begin{table}[ht]
\caption{\footnotesize The MSE of the location estimation based on the dataset gathered in the FIRST environment by comparing the ground truth (collected with the Vicon system) and the proposed $\RLF$.}\label{tab:table1}
\centering
\scalebox{.5}{\resizebox{\textwidth}{!}{\begin{tabular}{cc|c|c|c|c|} \hline
\multicolumn{1}{|c|}{\begin{tabular}[c]{@{}c@{}}\textbf{Movement Scenario}\end{tabular}}   & \textbf{AoA} & \textbf{RSSI}  &  \textbf{PDR} & \textbf{$\RLF$} \\ \hline
        \multicolumn{1}{|c|}{\begin{tabular}[c]{@{}c@{}}\textbf{Rectangular}\end{tabular}}      & 0.01979 &  0.12994 &  0.21997   & 0.003973\\ \hline
        \multicolumn{1}{|c|}{\begin{tabular}[c]{@{}c@{}}\textbf{Random}\end{tabular}} & 0.02508 &  0.14502 &  0.17303  & 0.00495\\ \hline
        \multicolumn{1}{|c|}{\begin{tabular}[c]{@{}c@{}}\textbf{Diagonal-A}\end{tabular}} & 0.0439 &  0.1961 &  0.2925  &  0.00685\\ \hline
        \multicolumn{1}{|c|}{\begin{tabular}[c]{@{}c@{}}\textbf{Diagonal-B}\end{tabular}} & 0.01992 &   0.16002 &  0.10029  & 0.00504\\ \hline
\end{tabular}}}
\end{table}
\begin{table}[ht]
\caption{\footnotesize The MSE of the location estimation based on the dataset gathered in the SECOND environment by comparing the ground truth (collected with the Vicon system) and the proposed $\RLF$.}\label{tab:table2}
\centering
\scalebox{.5}{\resizebox{\textwidth}{!}{\begin{tabular}{cc|c|c|c|c|} \hline
\multicolumn{1}{|c|}{\begin{tabular}[c]{@{}c@{}}\textbf{Movement Scenario}\end{tabular}}   & \textbf{AoA} & \textbf{RSSI}  &  \textbf{PDR} & \textbf{$\RLF$} \\ \hline
        \multicolumn{1}{|c|}{\begin{tabular}[c]{@{}c@{}}\textbf{Rectangular}\end{tabular}}      & 0.02193 &  0.05307 &  0.09464  &  0.006393\\ \hline
        \multicolumn{1}{|c|}{\begin{tabular}[c]{@{}c@{}}\textbf{Random}\end{tabular}} & 0.10133 &  0.09432 &  0.16103  & 0.00799\\ \hline
        \multicolumn{1}{|c|}{\begin{tabular}[c]{@{}c@{}}\textbf{Diagonal-A}\end{tabular}} & 0.11368 &  0.12331 &  0.19331  & 0.00994\\ \hline
        \multicolumn{1}{|c|}{\begin{tabular}[c]{@{}c@{}}\textbf{Diagonal-B}\end{tabular}} & 0.06854 &   0.11268 &  0.1105   & 0.00863\\ \hline
\end{tabular}}}
\end{table}
\begin{table}[ht]
\caption{\footnotesize The MSE of the location estimation based on the dataset gathered in the THIRD environment by comparing the ground truth (collected with the Vicon system) and the proposed $\RLF$.}\label{tab:table3}
\centering
\scalebox{.5}{\resizebox{\textwidth}{!}{\begin{tabular}{cc|c|c|c|c|} \hline
\multicolumn{1}{|c|}{\begin{tabular}[c]{@{}c@{}}\textbf{Movement Scenario}\end{tabular}}   & \textbf{AoA} & \textbf{RSSI}  &  \textbf{PDR} & \textbf{$\RLF$} \\ \hline
        \multicolumn{1}{|c|}{\begin{tabular}[c]{@{}c@{}}\textbf{Rectangular}\end{tabular}}      &  0.12379 &  1.63892 &  1.5869    & 0.0077 \\ \hline
        \multicolumn{1}{|c|}{\begin{tabular}[c]{@{}c@{}}\textbf{Random}\end{tabular}} & 0.02923 &  0.1123 &  0.14699  & 0.00745\\ \hline
        \multicolumn{1}{|c|}{\begin{tabular}[c]{@{}c@{}}\textbf{Diagonal-A}\end{tabular}} & 0.01468 &  0.22397 &  0.29195  &  0.00799\\ \hline
        \multicolumn{1}{|c|}{\begin{tabular}[c]{@{}c@{}}\textbf{Diagonal-B}\end{tabular}} & 0.01698 &   0.06556 &  0.0961  & 0.00648 \\ \hline
\end{tabular}}}
\end{table}
%
\begin{table}[ht]
\caption{\footnotesize Sample weight adjustment offered by the proposed $\RLF$, for information fusion of tracking scenarios.}\label{tab:table4-coefs}
\centering
\scalebox{.5}{\resizebox{\textwidth}{!}{\begin{tabular}{|cc|c|c|c|c|} \hline
\multicolumn{1}{|c|}{\begin{tabular}[c]{@{}c@{}}\textbf{Environment}\end{tabular}}  & \multicolumn{1}{|c|}{\begin{tabular}[c]{@{}c@{}}\textbf{Movement Scenario}\end{tabular}}   & \textbf{$\w^{\text{AoA}}$} & \textbf{$\w^{\text{RSSI}}$}  &  \textbf{$\w^{\text{PDR}}$} \\ \hline
\multirow{3}{4em}{\centering \textbf{First} Environment}   &   \multicolumn{1}{|c|}{\begin{tabular}[c]{@{}c@{}}\textbf{Rectangular}\end{tabular}}      &  0.1302776 &  0.7583678 &  0.1113545 \\
    &    \multicolumn{1}{|c|}{\begin{tabular}[c]{@{}c@{}}\textbf{Random}\end{tabular}} &  0.513865 &  0.825227 &  -0.339092 \\
     &   \multicolumn{1}{|c|}{\begin{tabular}[c]{@{}c@{}}\textbf{Diagonal-A}\end{tabular}} & 0.2742709 &  0.4277489 &  0.2979801  \\
    &    \multicolumn{1}{|c|}{\begin{tabular}[c]{@{}c@{}}\textbf{Diagonal-B}\end{tabular}} & 0.1022387 &   0.5218031 &  0.3759582 \\ \hline
\multirow{3}{4em}{\centering \textbf{Second} Environment}   &   \multicolumn{1}{|c|}{\begin{tabular}[c]{@{}c@{}}\textbf{Rectangular}\end{tabular}}      &  0.072684 &  1.0238779 &  -0.0965619 \\
    &    \multicolumn{1}{|c|}{\begin{tabular}[c]{@{}c@{}}\textbf{Random}\end{tabular}} & 0.1701491 &  0.6915486 &  0.1383023 \\
     &   \multicolumn{1}{|c|}{\begin{tabular}[c]{@{}c@{}}\textbf{Diagonal-A}\end{tabular}} & -0.3430738 &  0.734489 &  0.6085848 \\
    &    \multicolumn{1}{|c|}{\begin{tabular}[c]{@{}c@{}}\textbf{Diagonal-B}\end{tabular}} & 0.3344642 &   0.465767 &  0.1997688 \\  \hline
\multirow{3}{4em}{\centering \textbf{Third} Environment}   &   \multicolumn{1}{|c|}{\begin{tabular}[c]{@{}c@{}}\textbf{Rectangular}\end{tabular}}      &  -0.693659 &  1.0109855 &  0.6826736 \\
    &    \multicolumn{1}{|c|}{\begin{tabular}[c]{@{}c@{}}\textbf{Random}\end{tabular}} & 0.0188074 &  0.77122 &  0.2099726 \\
     &   \multicolumn{1}{|c|}{\begin{tabular}[c]{@{}c@{}}\textbf{Diagonal-A}\end{tabular}} & 0.0858805 &  0.5444015 &  0.3697179 \\
    &    \multicolumn{1}{|c|}{\begin{tabular}[c]{@{}c@{}}\textbf{Diagonal-B}\end{tabular}} & 0.2267579 &   0.7415241 &  0.0317179 \\  \hline
\end{tabular}}}
\end{table}
Table~\ref{tab:table1} shows the Mean Squared Error (MSE) (in meter) of the localization estimation in the first environment driven from the primary localization approaches including AoA, RSSI coupled with KF-PF and PDR, and also the proposed $\RLF$.
The same results can be seen in Table~\ref{tab:table2}, which compares the MSE of basic localization approaches and the proposed $\RLF$ in the second environment. As expected, the $\RLF$ shows the minimum MSE in location estimation compared to the baseline algorithms.
Table~\ref{tab:table3}, similar to the two previous tables, represents the MSE of the tracking approaches and their fusion strategies in the third environment. The proposed $\RLF$ outperforms the algorithms in this environment as well. According to the results, the $\RLF$ outperforms other tracking paths representing the high potentials of the proposed RL-based fusion solution.
One example of the results for the weights adjustment task performed via the proposed $\RLF$ framework can be seen in Table~\ref{tab:table4-coefs}. In this table, for each environment and all the moving scenarios, $\w^{\text{AoA}}$, $\w^{\text{RSSI}}$ and $\w^{\text{PDR}}$  are represented.
%
%

\subsection{Reliability of the Proposed $\RLF$ Framework}

\begin{figure}[t!]
\centering
\includegraphics[width=4.1cm]{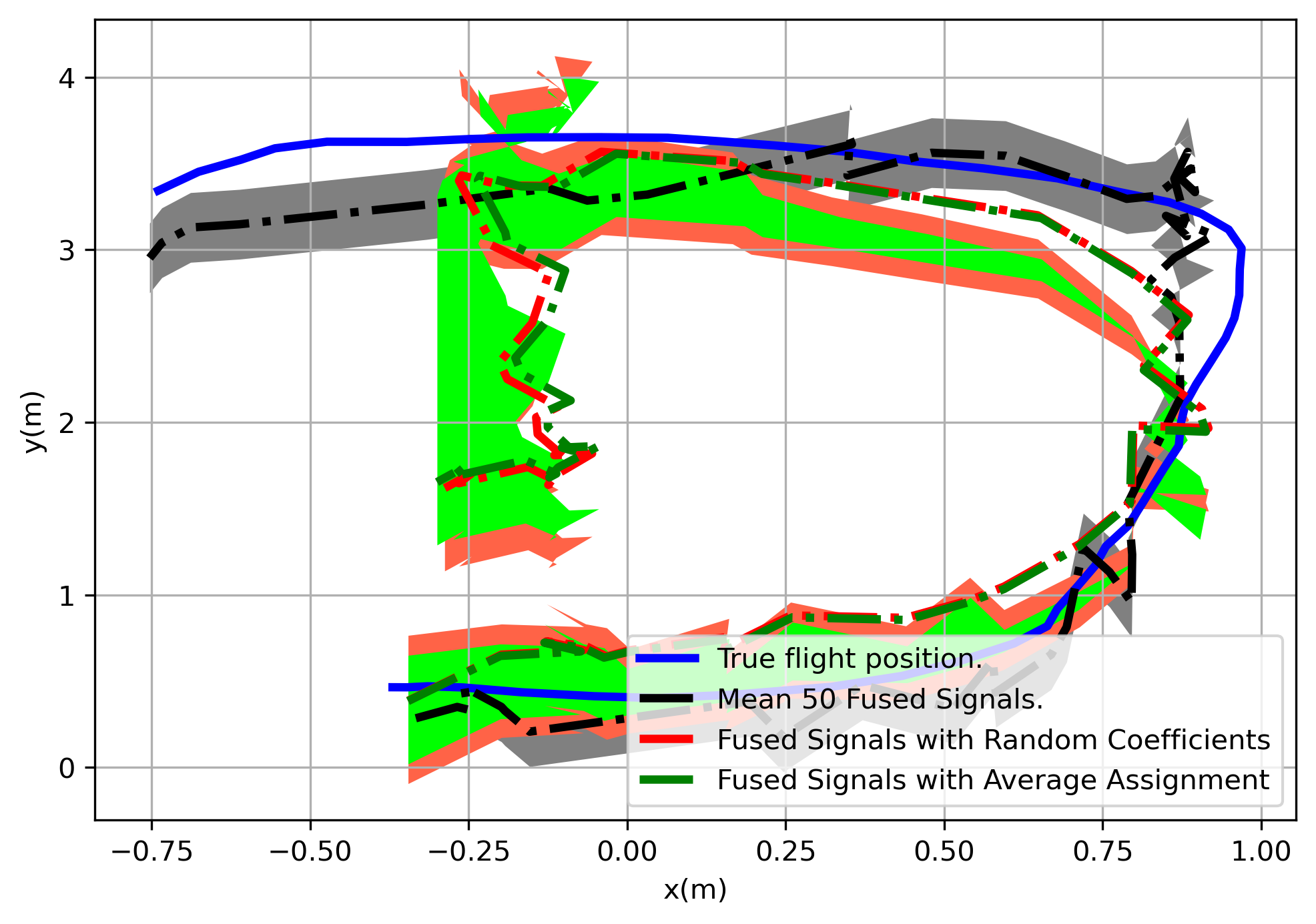}
\includegraphics[width=4.1cm]{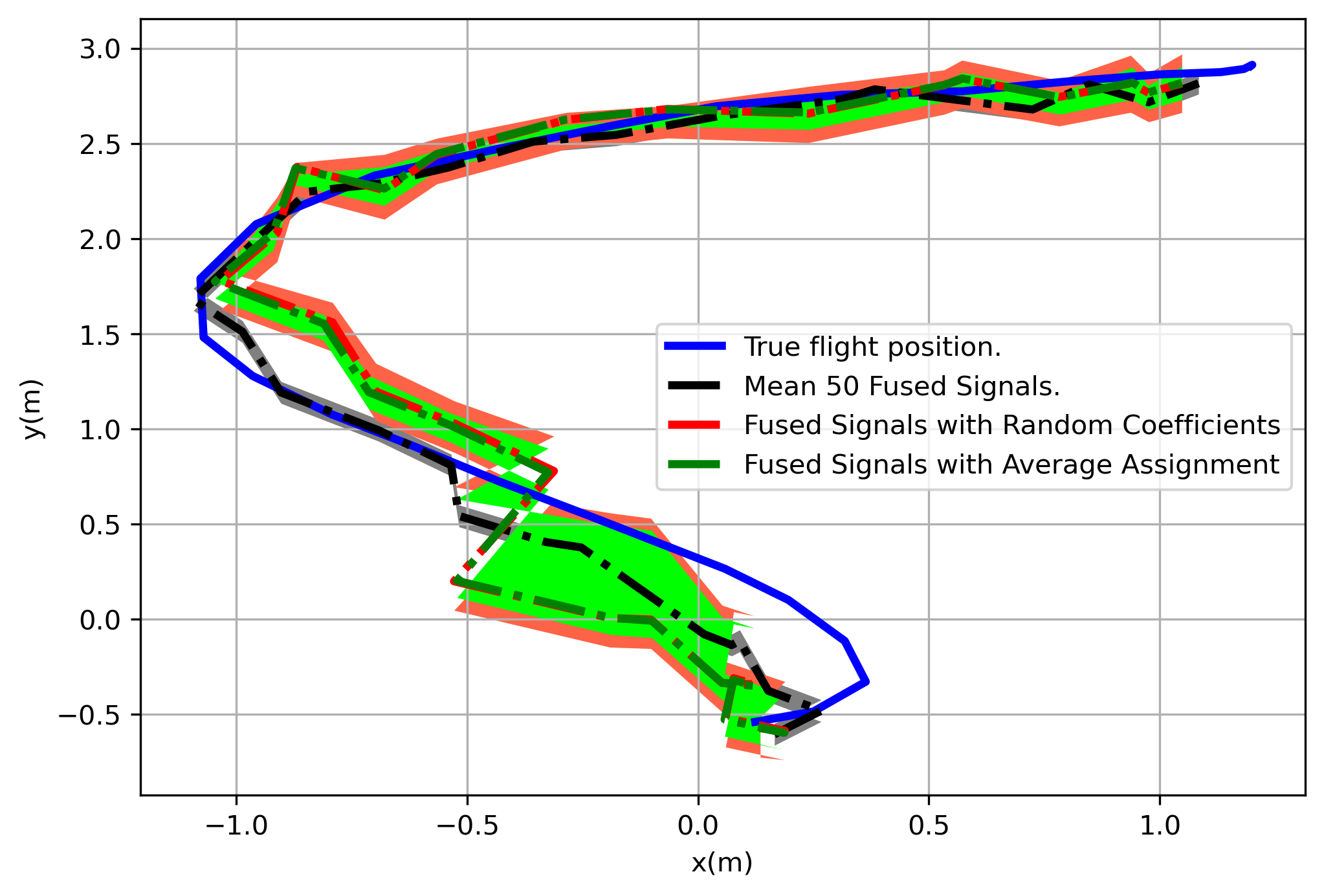}
\includegraphics[width=4.1cm]{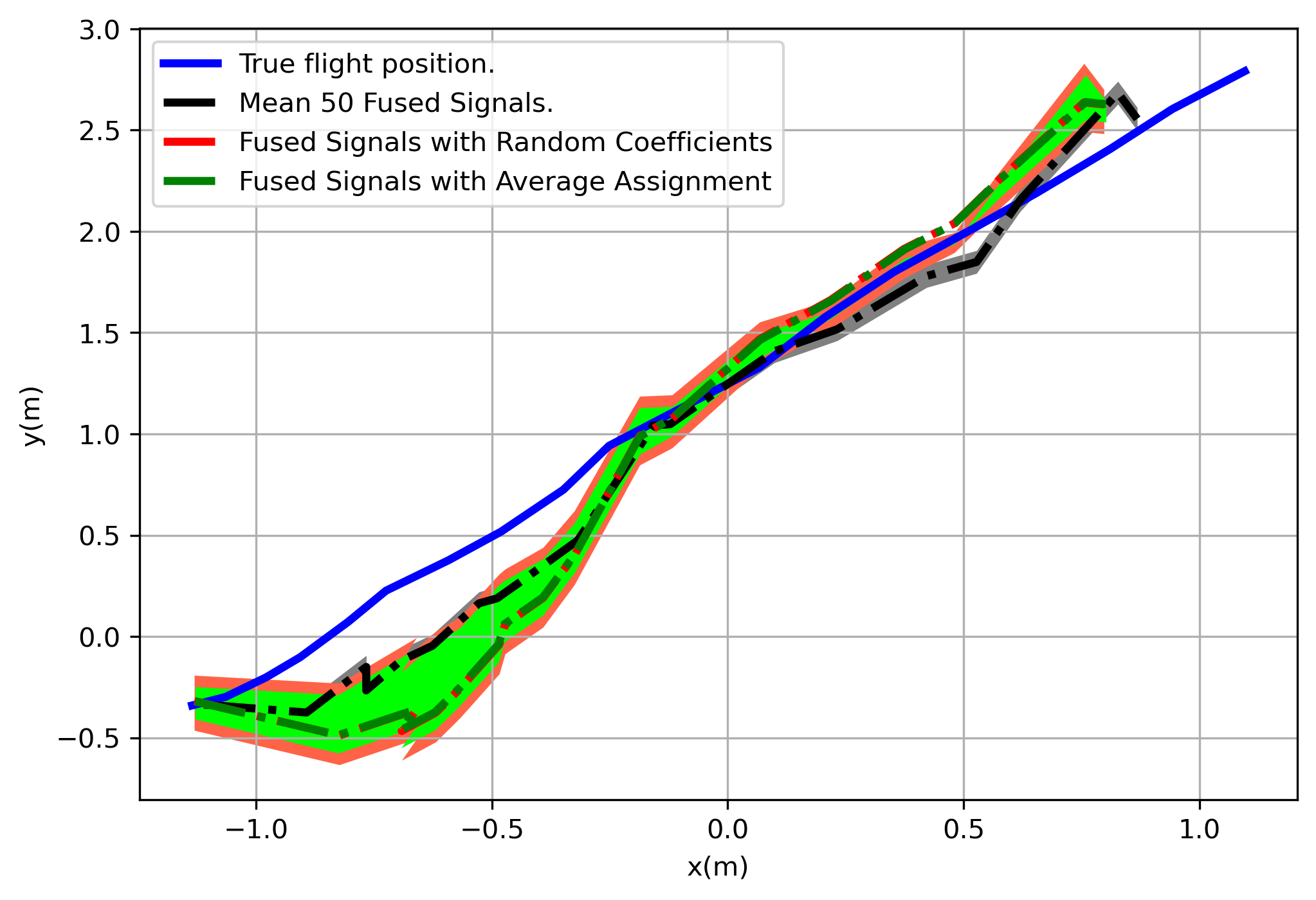}
\includegraphics[width=4.1cm]{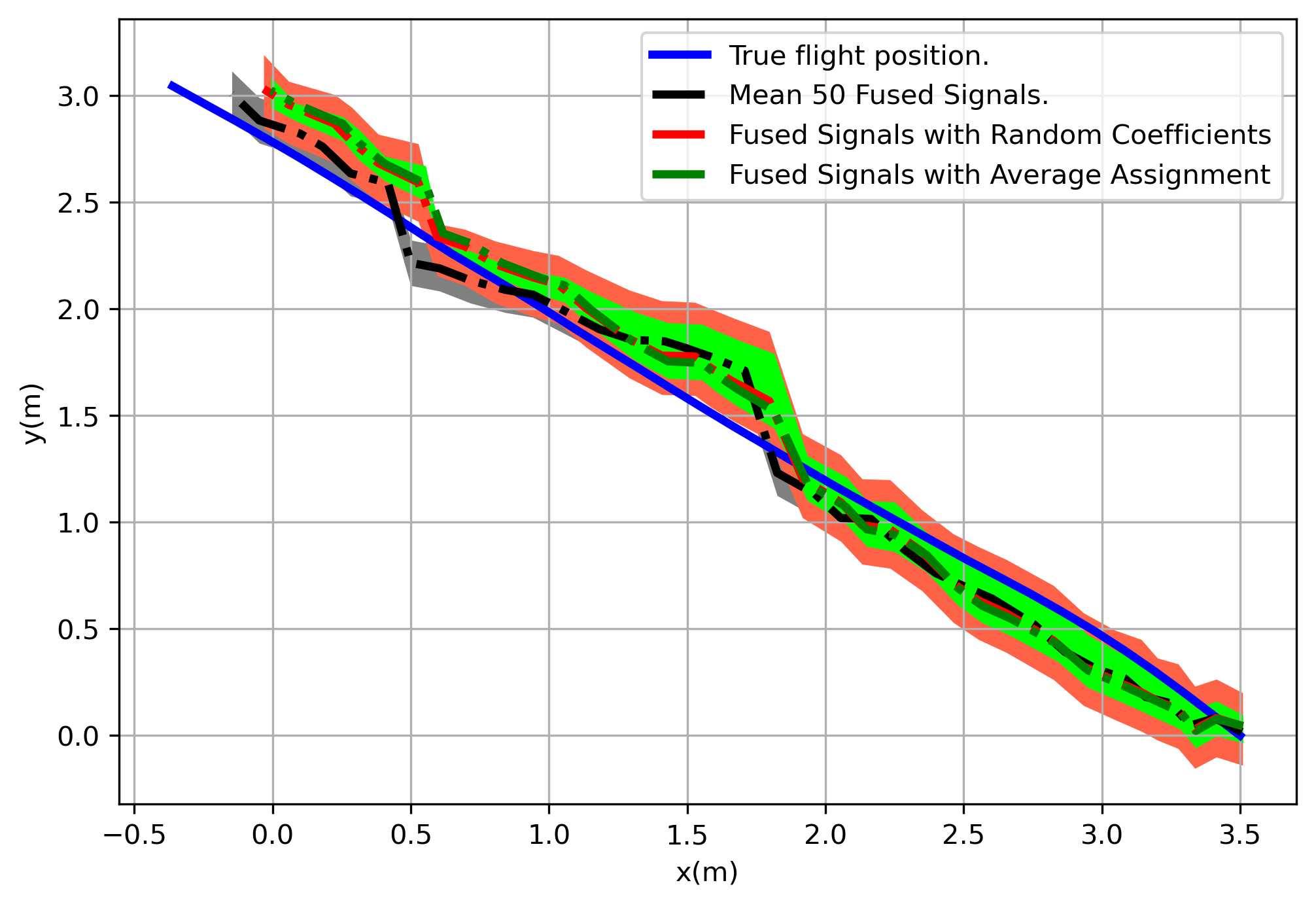}
\vspace{-.15in}
\caption{\footnotesize Evaluating the reliability of the proposed $\RLF$ compared to random weight adjustment and averaged weight selections schemas for four different sample movement scenarios: (a) Diagonal A;  (b) Diagonal B; (c) Random, and (d) Rectangular trajectories.}\label{Fig:reli25}
\vspace{.1in}
\end{figure}

\noindent
To verify  applicability of the proposed $\RLF$ in real-world scenarios, it is of significance to evaluate its reliability. A reliable learning procedure should be able to provide consistency in its performance and generate reproducible results over multiple runs of the model~\cite{Chan:2022, Salimibeni:2022}. Generally speaking, performance of RL-based solutions, particularly Deep Neural Network (DNN)-based approaches, are highly variable because of their dependence on a large number of tunable parameters. Hyperparameters, implementation details, and environmental factors are among these parameters~\cite{Salimibeni:2022}. This can result in unreliability of RL algorithms in real-world scenarios. To illustrate reliability of the proposed $\RLF$ framework, it is compared to other weight adjustment solutions. In one of these approaches, in each iteration, the weights are chosen randomly. In the other approach, all weights are set equal, and the fused tracking results is considered the averaged value of all the $3$ tracking paths, i.e., the RSSI path; the PDR path, and; the AoA path. More specifically, we have repeated each test $50$ times for all experiments. The RL fusion scenario consists of $1,000,000$ learning episodes together with $1,000$ test episodes.   As can be seen from Fig.~\ref{Fig:reli25}, for all four movement scenarios (i.e.,  Rectangular, Random, and two Diagonal), random weight adjustment and fixed averaged weights setting have higher variance illustrating their unreliability. Proposed $\RLF$ outperforms other approaches in terms of reliability and tracking performance . The ability to produce stable performance across different episodes is another aspect for investigating reliability of RL models. Stability of different models can also be compared through Fig.~\ref{Fig:reli25}. It can be seen that the proposed $\RLF$ algorithm is more stable than its counterparts as fewer sudden changes occur during different episodes.
%

\vspace{-.1in}
\section{Conclusion} \label{sec:6}
The paper presented $\RLF$, a RL-based information fusion framework by integrating BLE-based AoA,  PDR, and  RSSI approaches to improve overall tracking accuracy. The novelty of the proposed $\RLF$ framework lies in the integration of RL-based optimization for indoor tracking by leveraging different localization approaches. The proposed $\RLF$ framework is evaluated via several experiments across three different environments and four different moving scenarios including random movement which a challenging tracking task. For evaluation purposes, we looked at the MSE and reproducibility/stability aspects computed over multiple realizations. Based on the results, the proposed $\RLF$ framework outperformed its counterparts across different evaluation metrics.


\end{document}